\documentclass[11pt,a4paper]{article}
\usepackage[hyperref]{acl2020}
\usepackage{times}
\usepackage{latexsym}
\usepackage{multirow}
\usepackage{multicol}
\usepackage{graphicx}
\usepackage{tablefootnote}
\usepackage{threeparttable}
\usepackage{varwidth}
\usepackage{caption,newfloat}

\usepackage{hyperref}

\usepackage{microtype}

\aclfinalcopy 

\title{Simultaneous paraphrasing and translation by fine-tuning \\Transformer models}

\author{Rakesh Chada \\
  Amazon.com, Inc., \\
  \texttt{rakchada@amazon.com} \\
  }

\date{}

\begin{document}
\maketitle
\begin{abstract}
This paper describes the third place submission to the shared task on simultaneous translation and paraphrasing for language education at the 4th workshop on Neural Generation and Translation (WNGT) for ACL 2020. The final system leverages pre-trained translation models and uses a Transformer architecture combined with an oversampling strategy to achieve a competitive performance. This system significantly outperforms the baseline on Hungarian (27\% absolute improvement in Weighted Macro F1 score) and Portuguese (33\% absolute improvement) languages.
\end{abstract}

\section{Introduction}
This paper describes the third place submission to the shared task~\citet{staple20} on simultaneous translation and paraphrasing for language education at the 4th workshop on Neural Generation and Translation (WNGT) for ACL 2020. The shared task involves generating multiple translations for a given source text in English and a target language. The five target languages in the task are Hungarian (hu), Portuguese (pt), Japanese (ja), Korean (ko) and Vietnamese (vi). We competed in the Hungarian and Portuguese tracks. A goal of the shared task, hosted by Duolingo, is to enable development of automated grading processes and curation systems for language learners' responses.  A high-coverage and precise multi-output translation and paraphrasing system would vastly help such automated efforts. For the task, participants were provided with hand-crafted and field-tested sets of several possible translations for each English sentence. Each of these translations were also ranked and weighted according to actual learner response frequency and these weights were provided as additional features. Along with these, translations from AWS were provided as a baseline and additional data. The challenges associated with the shared task are two-fold: i) Translating from English to target languages and ii) Producing multiple valid translations (paraphrases) while balancing precision with the coverage. We conduct several experiments to address these two challenges and develop a simple system that leverages pre-trained transformer~\citet{vaswani2017attention} models and a wide beam search strategy. Furthermore, we leverage the provided translation scores and experiment with multiple training distribution strategies to develop a simple oversampling strategy that produces improvements over the vanilla method of using one translation one time. 

\section{Related work}
Paraphrasing and machine translation are well-studied research areas in general but there's not much research specifically in the context of multi-output translation systems, especially for low resource languages.~\citet{tan2019multilingual} train a Transformer-based Neural Machine Translation model for Hungarian-English and Portugese-English translation. However, their goal was to assess the benefits of multilingual modeling by clustering languages and is different from that of a multi-output translation system.  For English-Portuguese, ~\citet{aires-etal-2016-English} build a phrase-based machine translation system to translate biomedical texts. For multilingual parahrasing, ~\citet{ganitkevitch-callison-burch-2014-multilingual} release a database consisting of paraphrases for several languages, including Hungarian and Portuguese, at lexical, phrasal and syntactic level.~\citet{guo2019zeroshot} build a zero-shot multilingual paraphrase generation model to show mixed results. However, their end goal was to generate paraphrases in the same language (English) as opposed to our shared task which requires generating paraphrases in a different language.
\begin{table*}[!ht]
  \centering
  \begin{threeparttable}
  \renewcommand{\arraystretch}{1.2}
  \begin{tabular}{|p{3cm}|c|c|c|c|c|c|c|c|}
    \hline
    \multirow{2}{3cm}{\textbf{Target Language}}  & \multicolumn{6}{c|}{\textbf{Train}} & \multicolumn{1}{c|}{\textbf{Dev}} & \multicolumn{1}{c|}{\textbf{Test}}\\
    \cline{2-9}
    & \textbf{Prompts} & \textbf{Pairs} & \textbf{MSL}  & \textbf{MTL} & \textbf{99p SL}  & \textbf{99p TL} & \textbf{Prompts} & \textbf{Pairs}\\
    \hline
    Hungarian (hu) & 4000 & 251442 & 21 & 21 & 11 & 14 & 500 & 500 \\ \hline
    Portuguese (pt) & 4000 & 526466 & 33 & 21 & 25 & 15 & 500 & 500  \\ \hline
  \end{tabular}
  \caption{\label{stats-table} Dataset statistics. MSL=Maximum Source Length. MTL=Maximum Target Length. 99p SL=99th percentile Source Length. 99p TL=99th percentile Target Length.}
  \end{threeparttable}
\end{table*} 

\noindent~\citet{ippolito-etal-2019-comparison} study diverse decoding methods on conditional language models and show promising results on movie dialogue corpus and image captioning tasks.

\section{Task}
We describe dataset statistics and evaluation metrics in this section.

\subsection{Data}
\label{sect:data}
There are two phases of the competition - Dev and Test.
Table~\ref{stats-table} shows data statistics for all phases. There were 4000 train prompts provided, in English, for both Hungarian and Portuguese languages. However, each of these prompts were accompanied with multiple translations leading to 251,442 English-Hungarian (en-hu) pairs and 526,466 English-Portuguese (en-pt) pairs. There were 500 prompts in both dev and test phases. After tokenization, for en-hu, most of the source sentences were shorter than 11 tokens and target sentences were shorter than 14 tokens. For en-pt, most of the source sentences were shorter than 25 tokens and target sentences were shorter than 15 tokens.

\subsection{Evaluation Metrics}
\label{sect:evaluation}
The main scoring metric for the competition is the weighted macro F1 score. This is a measure of how well the system returns all human-curated translations weighted by the likelihood that an English learner would respond with each translation. For each prompt p, weighted macro F1 is calculated as the harmonic mean of precision and weighted recall (note that the precision is unweighted). To calculated weighted recall for each example, we first calculate Weighted True Positives (WTP) and Weighted False Negatives (WFN) as:
          \[WTP_p=\sum_{t \in TP_p} weight(t) \]
          \[WFN_p=\sum_{t \in FN_p} weight(t) \]
Then, weighted recall (WR) is calculated as:
\[WR_p=\frac{WTP_p}{WTP_p + WFN_p} \]
The weighted Macro F1 (WF) over all prompts P is then calculated by averaging over all prompts in the corpus as:
\[WF=\sum_{p \in P}\frac{WF_p}{|P|} \]

\section{System Design}
We now describe the final submitted system design in detail. We have experimented with several other variants and describe these in a later section~\ref{sect:ablation}.

\subsection{Data sampling}
\label{sect:sampling}
For the final system, we chose to use weighted sampling of the data where the weights correspond to the provided learner response frequency. Specifically, we multiply the frequency of the translation (a number between 0 and 1) with a heuristic value of 50 and duplicate the source-translation pair that many number of times. In effect, this would create repeated samples of certain pairs whose frequency is greater than 0.02 while eliminating pairs whose frequency is less than 0.02. With this sampling, we end up with 40,500 en-hu pairs and 42,000 en-pt pairs. We separate 15\% of the provided prompts as a validation set. The performance on this validation set is used to pick the best model.

\subsection{Preprocessing}
For text pre-processing, we use sentencepiece tokenization~\citet{kudo-richardson-2018-sentencepiece} for en-hu and byte-pair encoding~\citet{sennrich-etal-2016-neural} for en-pt data. We use pre-trained tokenization models provided in \href{https://github.com/Helsinki-NLP/OPUS-MT-train/tree/master/models}{OPUS-MT}.

\subsection{Model Architecture}

\begin{figure*}[!ht]
  \includegraphics[width=\linewidth]{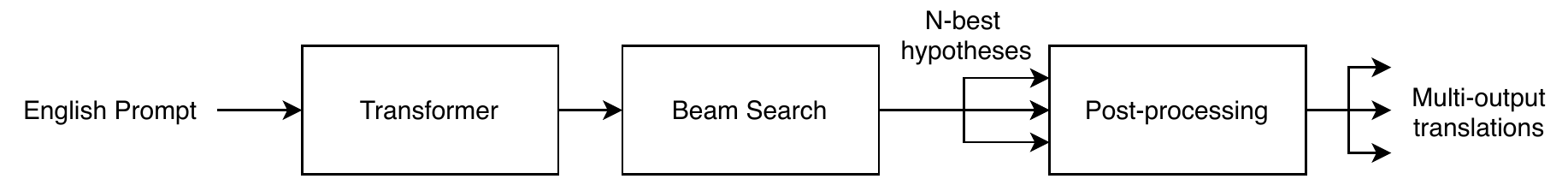}
  \caption{Architecture of the final system}
  \label{fig:duoarch}
\end{figure*}

The final submitted model architecture, shown in Figure~\ref{fig:duoarch}, uses the standard Transformer sequence-to-sequence model. This has 6 encoder and 6 decoder layers and an 8-headed attention mechanism in both encoder and decoder. We initialize the model with the pre-trained representations obtained from the \href{https://github.com/Helsinki-NLP/OPUS-MT-train/tree/master/models}{OPUS-MT data}. This model is then fine-tuned on the task data. We tie the encoder, decoder and output embedding weights and use a shared vocab size of 60,522. For position-wise feed-forward layers, the Swish activation function~\citet{ramachandran2018searching} is used. The whole model is fine-tuned, through an early stopping mechanism, on the dataset constructed as detailed in \ref{sect:sampling} .

\begin{table*}[!ht]
\center
\begin{threeparttable}
  \centering
  \renewcommand{\arraystretch}{1.2}
  \begin{tabular}{|p{3cm}|c|c|c|c|c|c|c|c|c|}
    \hline
    \multirow{2}{3cm}{\textbf{Model}} & \multicolumn{3}{c|}{\textbf{Validation}} & \multicolumn{3}{c|}{\textbf{Dev}} & \multicolumn{3}{c|}{\textbf{Test}}\\
    \cline{2-10}
    & \textbf{P} & \textbf{WR} & \textbf{WF} & \textbf{P} & \textbf{WR} & \textbf{WF} & \textbf{P} & \textbf{WR} & \textbf{WF}\\
    \hline
    Fairseq Baseline (en-hu) & - & - & - & 19.35 & 12.47 & 13.02 & 18.3 & 11.8 & 12.17 \\ \hline
    AWS Baseline (en-hu) & - & - & - & \textbf{84.6} & 19.9 & 29.85 & \textbf{86.8} & 18.9 & 28.1 \\ \hline
    Fine-tuned Transformer (en-hu) & 75.14 & 50.34 & 56.72 & 75.2 & \textbf{55.2} & \textbf{59.8} & 75.5 & \textbf{49.2} & \textbf{55.08}\\ \hline \hline
    Fairseq Baseline (en-pt) & - & - & - & 29.86 & 13.3 & 15.14 & 28.2 & 11.7 & 13.57 \\ \hline
    AWS Baseline (en-pt) & - & - & - & \textbf{86.8} & 14.09 & 21.15 & \textbf{87.8} & 13.9 & 21.3 \\ \hline
    Fine-tuned Transformer (en-pt) & 72.14 & 49.22 & 54.25 & 69.96 & \textbf{52.55} & \textbf{55.03} & 72.06 & \textbf{50.11} & \textbf{54.39}  \\ \hline
  \end{tabular}
  \caption{\label{results-table} Final submission results. \textbf{Bold} indicates best performance. P=Precision. WR=Weighted Recall. WF=Weighted Macro F1.}
  \end{threeparttable}
\end{table*}

\begin{table*}[!ht]
\center
\begin{threeparttable}
  \centering
  \renewcommand{\arraystretch}{1.2}
  \begin{tabular}{|p{3cm}|c|c|c|c|c|c|c|}
    \hline
    \multirow{2}{3cm}{\textbf{Model}} &  \multicolumn{7}{c|}{\textbf{Validation}}\\
    \cline{2-8}
    & \textbf{P} & \textbf{R} & \textbf{WR} & \textbf{MiF} & \textbf{MaF} & \textbf{WMiF} & \textbf{WMaF}\\
    \hline
    No fine-tuning & 52.41 & 6.32 & 41.18 & 11.28 & 19.21 & 46.12 & 41.31\\ \hline 
    No oversampling & 58.40 & 13.26 & 46.70 & \textbf{21.62} & \textbf{32.34} & 51.90 & 45.98\\ \hline
    No post-processing & 74.04 & 9.28 & 49.71 & 16.49 & 28.81 & 59.49 & 54.93\\ \hline
  \end{tabular}
  \caption{\label{ablations-table} Performance of various en-hu ablations on validation dataset. \textbf{Bold} indicates best performance. R=Recall. MiF=Micro F1. MaF=Macro F1. WMiF=Weighted Micro F1. WMaF=Weighted Macro F1.}
  \end{threeparttable}
\end{table*}

\begin{table*}[!ht]
\center
\begin{threeparttable}
  \centering
  \renewcommand{\arraystretch}{1.2}
  \begin{tabular}{|p{3cm}|c|c|c|c|c|c|c|}
    \hline
    \multirow{2}{3cm}{\textbf{Model}} &  \multicolumn{7}{c|}{\textbf{Validation}}\\
    \cline{2-8}
    & \textbf{P} & \textbf{R} & \textbf{WR} & \textbf{MiF} & \textbf{MaF} & \textbf{WMiF} & \textbf{WMaF}\\
    \hline
    Multi-output sequence & \textbf{74.44} & 7.33 & 44.35 & 13.35 & 23.58 & 55.59 & 52.29\\ \hline
    Nucleus sampling & 72.98 & 7.70 & 45.13 & 13.93 & 24.27 & 55.77 & 52.67\\ \hline
    Back Translation & 70.98 & 7.42 & 44.45 & 13.43 & 23.63 & 54.67 & 52.08\\ \hline
    Model-based Prediction Filtering & 72.71 & \textbf{10.60} & \textbf{51.90} & 18.51 & 31.01 & \textbf{60.56} & \textbf{56.10}\\ \hline
  \end{tabular}
  \caption{\label{variations-table} Performance of modeling variants on en-hu validation dataset. \textbf{Bold} indicates best performance.}
  \end{threeparttable}
\end{table*}

For fine-tuning, we use the standard cross-entropy loss objective on the target sequence along with a label smoothing loss~\citet{Szegedy2016RethinkingTI}.

For decoding, we use beam search with a beam size of 10 and select top 10 hypotheses for en-hu track. For en-pt track, we use a beam size of 28 and select top 28 hypotheses. We implement the model in Marian NMT~\citet{junczys-dowmunt-etal-2018-marian-fast}. 

\subsection{Postprocessing}
\label{sect: postprocessing}
The beam search outputs scores for each individual token. These scores represent the log likelihood of that token in the output sentence. As a post-processing step, we remove all translation predictions where the maximum of these token-level scores is less than -3.5. This value was determined by studying the impact of the maximum score thresholding on validation set performance.

\subsection{Hyperparameters}
We use the following hyperparameters. Batch size is set to 500. Dropout is set to 0.1. Label smoothing is set to 0.1. We use Adam optimizer with learning rate of 3e-4, $\beta_1$=0.9, $\beta_2$=0.98 and epsilon = 1e-9. We decay the learning rate by an inverse square root mechanism for 16000 steps. The gradient clip norm is set to 5. And patience for early stopping is set to 5.

\section{Ablations}
\label{sect:ablation}
\subsection{Ablations}
We have performed several ablation studies on the en-hu task. The results of all these studies are listed in Table~\ref{ablations-table}. We list the experiment methodologies below.\\
\textbf{No fine-tuning}: Here, we applied the pre-trained translation model directly on the task without any fine-tuning. The decoding was done using beam search beam size of 12  and by selecting top 12 hypotheses (determined based on validation performance).\\
\textbf{No oversampling}: Here, we use all provided translation pairs without any filtering based on the learner response frequency. We fine-tune the pre-trained model on this dataset and decode using beam search with a beam size of 15 and selecting top 15 hypotheses.\\
\textbf{No post-processing}: This is the same as the final submitted model without the post-processing (maximum score thresholding). 

\section{Other Modeling Variants}
\label{sect:variations}
We experimented with different modeling alternatives for the shared task. We describe them in this section. The results of these variations are listed in Table~\ref{variations-table}.
\subsection{Multi-output sequence formulation}
Here, we re-formulate the task as a multi-output prediction task by taking the top 5 translation pairs (based on the learner response frequency) and concatenating them into a single target sequence. The pre-trained model is then fine-tuned on this dataset. \\
\textbf{Nucleus sampling}: Here, we use the above multi-output sequence model and add Nucleus sampling~\citet{Holtzman2019TheCC} while decoding with p value set to 0.95.\\

\subsection{Back Translation}
Here, we start with a pre-trained hu-en translation model. We then construct a hu-en dataset from the provided en-hu translation pairs. The pre-trained model is fine-tuned on this dataset. We apply this fine-tuned hu-en model on the provided reference AWS translations of the target hu sentences. With a beam size of 15 and top-5 hypotheses selection, we generate 5 English paraphrases for each given English prompt. Now, the en-hu fine-tuned model from the ``Multi-output sequence formulation'' ablation is made to predict separately for each of the generated English paraphrases and all the outputs are combined into the final prediction.\\

\subsection{Model-based Prediction Filtering}
Here, we start with the final submission model and build a binary XGBoost classifier on top of it to filter predictions (accept vs reject). The features of the XGBoost model are the token-level scores, as described in Section~\ref{sect: postprocessing}, that are obtained from the final submission model. As different sequences have different lengths, we build a fixed size feature vector by truncating or padding all sequences to a length of 11. This is the 99 percentile source length listed in Table~\ref{stats-table}. The binary labels for training are obtained by comparing output translation with the provided gold translations. We do a randomized search on ``max\_depth'', ``colsample\_bytree'', ``colsample\_bylevel'' and ``n\_estimators'' hyperparameters of the XGBoost model to find the best set of values. We then perform a 5-fold cross-validation to identify the best model. The F1 score of this model on the ``accept'' class is 0.81 and on the ``reject'' class is 0.48. The overall accuracy is about 72\%.

\section{Results \& Discussion}
\label{sect:results}
Table~\ref{results-table} shows results of the final submission, for en-hu and en-pt tracks, along with a comparison to the baseline. As per the main evaluation metric (Weighted Macro F1 score), our model outperforms the strong AWS baseline by a significant margin on both en-hu and en-pt tracks. For en-hu, the improvement is about 30 absolute points on the dev dataset and 27 points on the test dataset. For en-pt, the improvement is about 34 absolute points on the dev dataset and 33 absolute points on the test dataset. This model ranked 1st on the dev leaderboard and 2nd on the test leaderboard for en-hu track. It ranked 2nd on the dev leaderboard and 3rd on the test leaderboard for en-pt track.\\

\noindent
Table~\ref{ablations-table} shows the results for several ablations for en-hu model listed in section~\ref{sect:ablation}. And Table~\ref{variations-table} shows results for several modeling variants listed in section~\ref{sect:variations}.
There are several interesting observations to be made from these ablations and variants. First, there's a clear improvement of about 15.4 points in Weighted Macro F1 from fine-tuning the pre-trained model on the provided dataset. The simple post-processing strategy of score thresholding yielded a gain of about 1.79 absolute points. Similarly, there's also a big improvement of about 10.7 absolute points from the oversampling strategy we used (as opposed to no oversampling). However, this gap seemed to have been closed by a big margin (about 7 absolute points) through the multi-output sequence formulation and slightly more by adding Nucleus sampling on top of it. A separate approach that uses back translation seemed to also have yielded similar gains upon the ``No oversampling'' approach. The model-based prediction filtering yielded an improvement of about 4 absolute points.  Interestingly, all of these variants still ended up inferior (by varying levels) to the simple oversampling + fine-tuning + post-processing strategy that was used for the final submission.

\section{Summary}
We describe the system for our submission to the shared task on simultaneous translation and paraphrasing for language education at the 4th workshop on Neural Generation and Translation (WNGT) for ACL 2020. The final submitted system leverages pre-trained translation models, with Transformer architecture, and an oversampling strategy to achieve competitive performance. For future, it'd be interesting to see if initializing the model with latest state-of-the-art sequence-to-sequence pre-trained models such as BART~\citet{lewis2019bart} and T5~\citet{raffel2019exploring} and fine-tuning could help boost performance. It would also be a promising direction to explore the benefit of using cross-lingual models such as XLM-Roberta~\citet{conneau2019unsupervised}. One way to use them would be to initialize the encoder part of the architecture with pre-trained representations. Given the shared representations, it might be interesting to see if concatenating several language pairs' train datasets and training a joint model produces additional benefits.

\section*{Acknowledgments}

We thank the Duolingo team for providing the dataset and organizing the competition and thank reviewers for providing valuable feedback.

\bibliography{anthology,acl2020}

\begin{thebibliography}{16}
\expandafter\ifx\csname natexlab\endcsname\relax\def\natexlab#1{#1}\fi

\bibitem[{Aires et~al.(2016)Aires, Lopes, and Gomes}]{aires-etal-2016-English}
Jos{\'e} Aires, Gabriel Lopes, and Lu{\'\i}s Gomes. 2016.
\newblock \href {https://doi.org/10.18653/v1/W16-2335} {{E}nglish-{P}ortuguese
  biomedical translation task using a genuine phrase-based statistical machine
  translation approach}.
\newblock In \emph{Proceedings of the First Conference on Machine Translation:
  Volume 2, Shared Task Papers}, pages 456--462, Berlin, Germany. Association
  for Computational Linguistics.

\bibitem[{Conneau et~al.(2019)Conneau, Khandelwal, Goyal, Chaudhary, Wenzek,
  Guzmán, Grave, Ott, Zettlemoyer, and Stoyanov}]{conneau2019unsupervised}
Alexis Conneau, Kartikay Khandelwal, Naman Goyal, Vishrav Chaudhary, Guillaume
  Wenzek, Francisco Guzmán, Edouard Grave, Myle Ott, Luke Zettlemoyer, and
  Veselin Stoyanov. 2019.
\newblock \href {http://arxiv.org/abs/1911.02116} {Unsupervised cross-lingual
  representation learning at scale}.

\bibitem[{Ganitkevitch and
  Callison-Burch(2014)}]{ganitkevitch-callison-burch-2014-multilingual}
Juri Ganitkevitch and Chris Callison-Burch. 2014.
\newblock \href
  {http://www.lrec-conf.org/proceedings/lrec2014/pdf/659_Paper.pdf} {The
  multilingual paraphrase database}.
\newblock In \emph{Proceedings of the Ninth International Conference on
  Language Resources and Evaluation ({LREC}-2014)}, pages 4276--4283,
  Reykjavik, Iceland. European Languages Resources Association (ELRA).

\bibitem[{Guo et~al.(2019)Guo, Liao, Jiang, Zhang, Zhang, and
  Liu}]{guo2019zeroshot}
Yinpeng Guo, Yi~Liao, Xin Jiang, Qing Zhang, Yibo Zhang, and Qun Liu. 2019.
\newblock \href {http://arxiv.org/abs/1911.03597} {Zero-shot paraphrase
  generation with multilingual language models}.

\bibitem[{Holtzman et~al.(2019)Holtzman, Buys, Forbes, and
  Choi}]{Holtzman2019TheCC}
Ari Holtzman, Jan Buys, Maxwell Forbes, and Yejin Choi. 2019.
\newblock The curious case of neural text degeneration.
\newblock \emph{ArXiv}, abs/1904.09751.

\bibitem[{Ippolito et~al.(2019)Ippolito, Kriz, Sedoc, Kustikova, and
  Callison-Burch}]{ippolito-etal-2019-comparison}
Daphne Ippolito, Reno Kriz, Joao Sedoc, Maria Kustikova, and Chris
  Callison-Burch. 2019.
\newblock \href {https://doi.org/10.18653/v1/P19-1365} {Comparison of diverse
  decoding methods from conditional language models}.
\newblock In \emph{Proceedings of the 57th Annual Meeting of the Association
  for Computational Linguistics}, pages 3752--3762, Florence, Italy.
  Association for Computational Linguistics.

\bibitem[{Junczys-Dowmunt et~al.(2018)Junczys-Dowmunt, Grundkiewicz, Dwojak,
  Hoang, Heafield, Neckermann, Seide, Germann, Aji, Bogoychev, Martins, and
  Birch}]{junczys-dowmunt-etal-2018-marian-fast}
Marcin Junczys-Dowmunt, Roman Grundkiewicz, Tomasz Dwojak, Hieu Hoang, Kenneth
  Heafield, Tom Neckermann, Frank Seide, Ulrich Germann, Alham~Fikri Aji,
  Nikolay Bogoychev, Andr{\'e} F.~T. Martins, and Alexandra Birch. 2018.
\newblock \href {https://doi.org/10.18653/v1/P18-4020} {{M}arian: Fast neural
  machine translation in {C}++}.
\newblock In \emph{Proceedings of {ACL} 2018, System Demonstrations}, pages
  116--121, Melbourne, Australia. Association for Computational Linguistics.

\bibitem[{Kudo and Richardson(2018)}]{kudo-richardson-2018-sentencepiece}
Taku Kudo and John Richardson. 2018.
\newblock \href {https://doi.org/10.18653/v1/D18-2012} {{S}entence{P}iece: A
  simple and language independent subword tokenizer and detokenizer for neural
  text processing}.
\newblock In \emph{Proceedings of the 2018 Conference on Empirical Methods in
  Natural Language Processing: System Demonstrations}, pages 66--71, Brussels,
  Belgium. Association for Computational Linguistics.

\bibitem[{Lewis et~al.(2019)Lewis, Liu, Goyal, Ghazvininejad, Mohamed, Levy,
  Stoyanov, and Zettlemoyer}]{lewis2019bart}
Mike Lewis, Yinhan Liu, Naman Goyal, Marjan Ghazvininejad, Abdelrahman Mohamed,
  Omer Levy, Ves Stoyanov, and Luke Zettlemoyer. 2019.
\newblock \href {http://arxiv.org/abs/1910.13461} {Bart: Denoising
  sequence-to-sequence pre-training for natural language generation,
  translation, and comprehension}.

\bibitem[{Mayhew et~al.(2020)Mayhew, Bicknell, Brust, McDowell, Monroe, and
  Settles}]{staple20}
Stephen Mayhew, Klinton Bicknell, Chris Brust, Bill McDowell, Will Monroe, and
  Burr Settles. 2020.
\newblock Simultaneous translation and paraphrase for language education.
\newblock In \emph{Proceedings of the ACL Workshop on Neural Generation and
  Translation (WNGT)}. ACL.

\bibitem[{Raffel et~al.(2019)Raffel, Shazeer, Roberts, Lee, Narang, Matena,
  Zhou, Li, and Liu}]{raffel2019exploring}
Colin Raffel, Noam Shazeer, Adam Roberts, Katherine Lee, Sharan Narang, Michael
  Matena, Yanqi Zhou, Wei Li, and Peter~J. Liu. 2019.
\newblock \href {http://arxiv.org/abs/1910.10683} {Exploring the limits of
  transfer learning with a unified text-to-text transformer}.

\bibitem[{Ramachandran et~al.(2018)Ramachandran, Zoph, and
  Le}]{ramachandran2018searching}
Prajit Ramachandran, Barret Zoph, and Quoc~V. Le. 2018.
\newblock \href {https://openreview.net/forum?id=SkBYYyZRZ} {Searching for
  activation functions}.

\bibitem[{Sennrich et~al.(2016)Sennrich, Haddow, and
  Birch}]{sennrich-etal-2016-neural}
Rico Sennrich, Barry Haddow, and Alexandra Birch. 2016.
\newblock \href {https://doi.org/10.18653/v1/P16-1162} {Neural machine
  translation of rare words with subword units}.
\newblock In \emph{Proceedings of the 54th Annual Meeting of the Association
  for Computational Linguistics (Volume 1: Long Papers)}, pages 1715--1725,
  Berlin, Germany. Association for Computational Linguistics.

\bibitem[{Szegedy et~al.(2016)Szegedy, Vanhoucke, Ioffe, Shlens, and
  Wojna}]{Szegedy2016RethinkingTI}
Christian Szegedy, Vincent Vanhoucke, Sergey Ioffe, Jon Shlens, and Zbigniew
  Wojna. 2016.
\newblock Rethinking the inception architecture for computer vision.
\newblock \emph{2016 IEEE Conference on Computer Vision and Pattern Recognition
  (CVPR)}, pages 2818--2826.

\bibitem[{Tan et~al.(2019)Tan, Chen, He, Xia, Qin, and
  Liu}]{tan2019multilingual}
Xu~Tan, Jiale Chen, Di~He, Yingce Xia, Tao Qin, and Tie-Yan Liu. 2019.
\newblock \href {http://arxiv.org/abs/1908.09324} {Multilingual neural machine
  translation with language clustering}.

\bibitem[{Vaswani et~al.(2017)Vaswani, Shazeer, Parmar, Uszkoreit, Jones,
  Gomez, Kaiser, and Polosukhin}]{vaswani2017attention}
Ashish Vaswani, Noam Shazeer, Niki Parmar, Jakob Uszkoreit, Llion Jones,
  Aidan~N Gomez, {\L}ukasz Kaiser, and Illia Polosukhin. 2017.
\newblock Attention is all you need.
\newblock In \emph{Advances in neural information processing systems}, pages
  5998--6008.

\end{thebibliography}
\bibliographystyle{acl_natbib}

\end{document}